%% file: main.tex
\documentclass[lettersize,journal]{IEEEtran}
\usepackage{amsmath,amsfonts}
\usepackage{algorithmic}
\usepackage{algorithm}
\usepackage{array}
\usepackage[caption=true,font=normalsize,labelfont=sf,textfont=sf]{subfig}
\usepackage{textcomp}
\usepackage{stfloats}
\usepackage{url}
\usepackage{verbatim}
\usepackage{cite}
\usepackage{graphicx}
\usepackage{booktabs}
\usepackage{multirow}
\usepackage{wrapfig}
\usepackage{xcolor}
\usepackage{orcidlink}
\usepackage{etoolbox}
\usepackage{colortbl}
\usepackage{amsmath, amssymb, algorithm, algorithmic}
\newcommand{\ie}{\emph{i.e.\ }}

\newcommand{\etal}{\emph{et al.}}
\AtEndPreamble{
    \usepackage[capitalize]{cleveref}
    \crefname{section}{Sec.}{Secs.}
    \Crefname{section}{Section}{Sections}
    \crefname{table}{Tab.}{Tabs.}
    \Crefname{table}{Table}{Tables}
}
\begin{document}

\title{Mitigating Low-Frequency Bias: Feature Recalibration and Frequency Attention Regularization for Adversarial Robustness}

\author{Kejia~Zhang$^{\orcidlink{0000-0002-7303-1428}}$, 
        Juanjuan~Weng$^{\orcidlink{0000-0003-0825-2272}}$,
        Yuanzheng~Cai, 
        Zhiming~Luo$^{\orcidlink{0000-0002-3411-9582}}$, \IEEEmembership{Member, IEEE},
        Shaozi~Li$^{\orcidlink{0000-0001-5403-9945}}$
\thanks{This work is supported by the National Natural Science Foundation of China (No.~62276221, No.~62376232). 
\it{(Corresponding author: Zhiming Luo and Juanjuan Weng.)}
}
\IEEEcompsocitemizethanks{\IEEEcompsocthanksitem Kejia Zhang, Zhiming Luo, and Shaozi Li are with the Department of Artificial Intelligence, Xiamen University, Xiamen 361005, China.
\IEEEcompsocthanksitem Yuanzheng Cai is with Fujian Provincial Key Laboratory of Information Processing and Intelligent Control, College of Computer and Control Engineering, Minjiang University, Fuzhou, 350108, China.
\IEEEcompsocthanksitem Juanjuan Weng is with the College of Information Science and Technology, Jinan University, Guangzhou, 510632, China.
}
}
% .

% The paper headers
\markboth{}%
{Shell \MakeLowercase{\textit{et al.}}: A Sample Article Using IEEEtran.cls for IEEE Journals}

\IEEEpubid{}
% Remember, if you use this you must call \IEEEpubidadjcol in the second
% column for its text to clear the IEEEpubid mark.

\maketitle

\begin{abstract}
Ensuring the robustness of deep neural networks against adversarial attacks remains a fundamental challenge in computer vision. While adversarial training (AT) has emerged as a promising defense strategy, our analysis reveals a critical limitation: AT-trained models exhibit a bias toward low-frequency features while neglecting high-frequency components. This bias is particularly concerning as each frequency component carries distinct and crucial information: low-frequency features encode fundamental structural patterns, while high-frequency features capture intricate details and textures. To address this limitation, we propose High-Frequency Feature Disentanglement and Recalibration (HFDR), a novel module that strategically separates and recalibrates frequency-specific features to capture latent semantic cues. We further introduce frequency attention regularization to harmonize feature extraction across the frequency spectrum and mitigate the inherent low-frequency bias of AT. Extensive experiments demonstrate our method's superior performance against white-box attacks and transfer attacks, while exhibiting strong generalization capabilities across diverse scenarios.
\end{abstract}

\begin{IEEEkeywords}
Neural Network Robustness, Adversarial Training, Frequency.
\end{IEEEkeywords}

\section{Introduction}
    DNNs have achieved remarkable success in various applications across diverse domains.
    However, their susceptibility to inconspicuous adversarial perturbations~\cite{goodfellow2014explaining, zhu2023information, jia2022adv} remains a substantial concern. These subtle perturbations, which are undetectable by the human eye, have the potential threat of resulting in erroneous predictions~\cite{perturbation,  oh2017adversarial}. 
    As a consequence, the security and robustness of DNNs have become areas of widespread concern within both academic and industrial communities. 
    Adversarial training (AT)~\cite{jia2022adversarial, andriushchenko2020understanding, kuang2024defense, jin2023randomized} is widely acknowledged as one of the most effective techniques to enhance adversarial robustness of models. 
    Unlike normal training (NT), which uses clean data to minimize the standard classification loss, AT further incorporates adversarial examples (AEs)—carefully crafted perturbations designed to exploit model vulnerabilities—into the training process.
    \par 
    Despite AT's remarkable performance, relying solely on AE-based data augmentation proves insufficient as a comprehensive solution.
    Research has extensively explored regularization~\cite{Regular_1, Regular_2} and feature denoising~\cite{Denoising_1, Denoising_2} methods to enhance adversarial robustness. 
    Recent Fourier analysis-based advances~\cite{2019_NIPS_Fourier,2020_CVPR_Fourier,2022_CVPR_Fourier,wang2024boosting} highlight the benefits of incorporating frequency analysis during AT to improve robustness.
    Building on these insights, Bu \etal ~\cite{2023_ICCV_frequency} developed an adaptive module to control frequency preferences by adjusting low and high-frequency components in feature representations. Similarly, Yucel \etal ~\cite{Yucel_2023_ICCV} introduced HybridAugment, a data augmentation method designed to reduce CNNs' reliance on high-frequency components, followed by HybridAugment++ to unify various frequency-spectrum augmentations.
    However, the tendency of these approaches to discard or suppress high-frequency components may result in a significant loss of semantic information~\cite{2020_CVPR_Fourier}.
    
    \begin{figure}[t]
        \begin{center}
        \includegraphics[width=\linewidth]{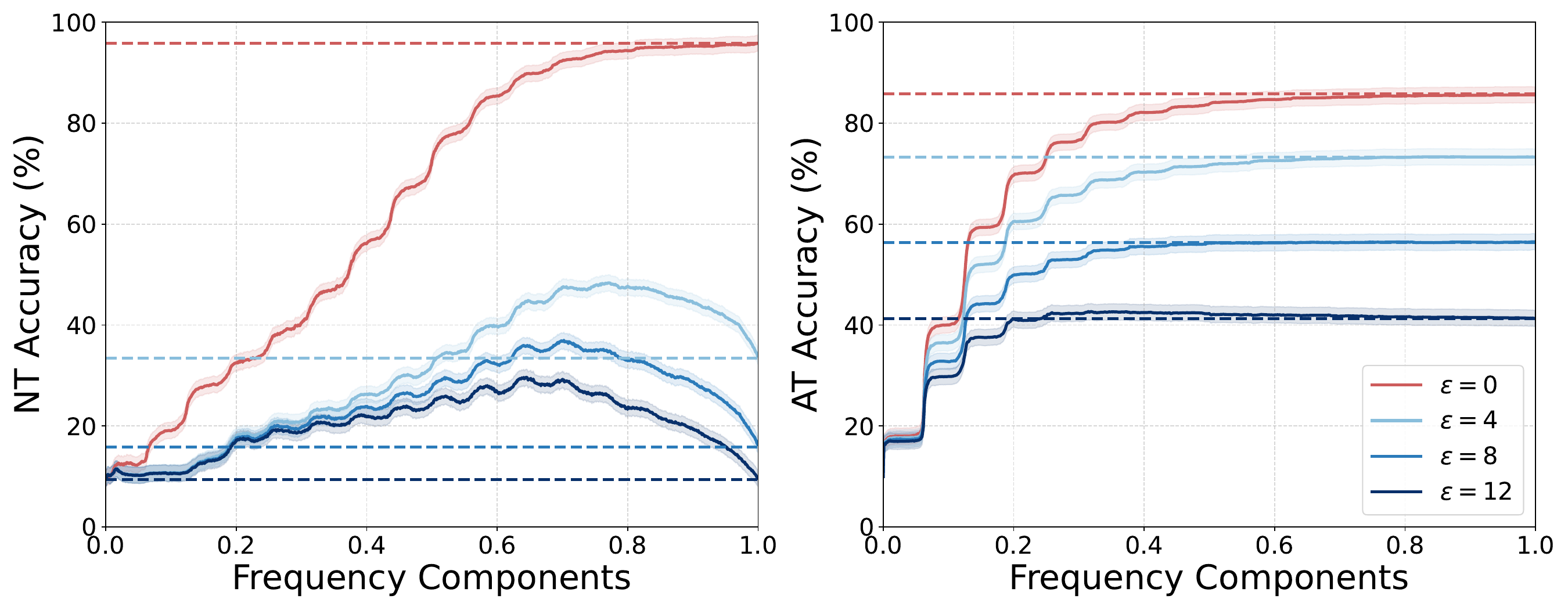}
        \end{center}
        \caption{
        Analysis of model performance across frequency components under different adversarial perturbation strengths ($\epsilon$ = 0, 4, 8, 12) for normal training (NT, left) and adversarial training (AT, right). The x-axis shows the percentage of frequency components retained after Fourier transformation.
        }
        \label{Model_robust_freq}
    \end{figure}
    
\begin{figure*}[t]
        \begin{center}
        \includegraphics[width=\linewidth]{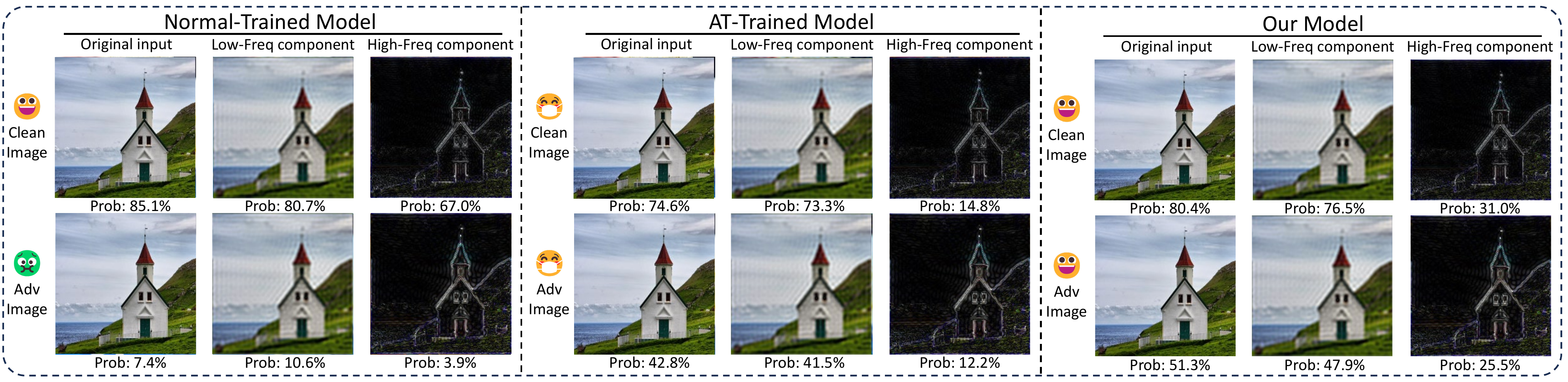}
        \end{center}
        \caption{
        Predicted confidence (post-softmax probability) comparison across different frequency components for clean and adversarial (Adv) images. Results shown for normal-trained, adversarial-trained, and our proposed model. The original inputs are decomposed into high and low-frequency components via 2D Discrete Fourier Transform (DFT).
        }
        \label{High_low_prob}
    \end{figure*}

    We present a comprehensive frequency-domain analysis of adversarial robustness through two experiments illustrated in \Cref{Model_robust_freq} and \Cref{High_low_prob}. Our analysis reveals fundamental differences between normally-trained (NT) and adversarially-trained (AT) models in their utilization of frequency components.

    In \Cref{Model_robust_freq}, 
    we analyze the impact of retaining different frequency components on model performance under varying adversarial perturbation strengths. The results reveal distinct behaviors between NT and AT models:
    \textbf{(1) For NT models,}
    performance on clean images ($\epsilon=0$) improves as more frequency components are included, highlighting the significance of all frequencies in the classification of clean images. 
    However, under adversarial attack conditions ($\epsilon=4,8,12$), the model's performance initially improves with an increase in frequency components, but degrades significantly as more high-frequency features are retained. These results underscore that adversarial attacks primarily affect the high-frequency components, thus affecting the classification accuracy. Notably, the accuracies of the NT model decrease sharply as the strength of adversarial perturbations increases, further emphasizing its sensitivity to attack intensity.
    \textbf{(2) For AT models,} they exhibit a different characteristic. 
    While their accuracy on clean images is slightly lower than that of NT models, they demonstrate superior robustness under adversarial perturbations. The AT model also experiences performance degradation under adversarial attacks, but the decline is significantly less pronounced than that of the NT model. 
    Furthermore, the classification accuracies of AT models quickly reached saturation, and the inclusion of more high-frequencies has minimal impact on the performance. 
    These findings indicate that the AT model heavily depends on low-frequency components for classification while overlooking high-frequency information. Consequently, adversarial training induces a low-frequency bias.
    
    In \Cref{High_low_prob}, we further illustrate the predicted confidence across different frequency components. \textbf{(1) NT Model:} Both high-frequency and low-frequency components from the clean image obtain high post-softmax probabilities. Besides, the original input gets the highest probability, indicating both high-frequency and low-frequency are important for the final probability. For adversarial images, the post-softmax probabilities are significantly dropped across all components, with high-frequency features being more severely impacted, revealing their vulnerability to adversarial perturbations.
    \textbf{(2) AT Model:} The AT-trained model relies predominantly on low-frequency components for classification in both clean and adversarial images. High-frequency components exhibit notably lower post-softmax probabilities, suggesting that AT-trained models overlook the high-frequency components for the final classification.
    \textbf{(3)}
    Comparing the results between NT and AT models, we can find that adversarial training tends to rely on the low-frequency components for the classification, while neglecting the information in the high frequencies of both clean image and adversarial images.
    This observation is consistent with the low-frequency bias in Figure~\ref{Model_robust_freq}, where the AT models fail to utilize high-frequency features effectively.
    \par 
    Our analysis reveals a fundamental contradiction in how adversarially trained networks process frequency-dependent features. Low-frequency feature robustness manifests as the network's capacity to effectively capture semantically valuable information for predictions,  despite perturbations in the low-frequency domain. Conversely, high-frequency feature robustness reflects the network's diminished sensitivity to fine-grained details, potentially compromising its ability to capture critical textural nuances ~\cite{oechsle2019texture}.
    \par
    Based on the above-discussed valuable insights, we raise a hypothesis: is it possible to improve the model's robustness by utilizing high-frequency characteristics that undergo high-intensity perturbations? 
    This entails capturing latent cues from high-frequency features and harmonizing the network's feature extraction capabilities across various frequency domains. The objective of this approach is to recalibrate the high-frequency features and mitigate the low-frequency bias caused by AT.
    \par 
    To achieve this, we proposed a novel module ``High-Frequency Disentanglement and Recalibration (HFDR)''. 
    Firstly, HFDR employs high-pass filters to generate element-wise frequency attention maps, facilitating the disentanglement of features across different frequency domains. Then through the recalibration of vulnerable high-frequency features, the model effectively captures potential semantic cues within the high-frequency domain. The fusion of recalibrated high-frequency features with low-frequency features enhances the network's ability to extract distinct frequency features. To further mitigate the low-frequency biases induced by AT, we introduce a frequency domain attention regularization. This regularization strategy aims to harmonize the network's extraction capacity for different frequency domain information.
    Notably, our proposed model achieves comparable or superior robustness compared to state-of-the-art methods, while incurring minimal additional computational overhead. Addtionally, the results of our model in Figure~\ref{High_low_prob} show a significant increase in post-softmax probabilities associated with the high-frequency components.
    \par 
    In summary, this study makes the following three key contributions:
    \begin{itemize}
        \item This study identifies a significant low-frequency bias in conventional adversarial training methods, which often overlook high-frequency features containing valuable semantic information.
        \item We propose a High-Frequency Disentanglement and Recalibration (HFDR) module that extracts latent high-frequency cues, integrates frequency attention mechanisms, and addresses low-frequency bias to improve model robustness.
        \item The proposed model achieves state-of-the-art robustness with minimal additional computational cost and can be seamlessly integrated with existing adversarial training methods to further enhance model performance.
    \end{itemize}

\section{Related Work}
\subsection{Adversarial Attack}
\label{attack section}
Deep neural networks (DNNs) demonstrate remarkable performance in various tasks by learning intricate relationships between inputs and outputs through complex, non-linear, and high-dimensional mappings. 
However, DNNs are susceptible to adversarial attacks, in which carefully crafted perturbation can deceive the model, leading to incorrect predictions.
Several research studies have been dedicated to investigating the vulnerability of models and have proposed various advanced adversarial attack methods.
Cao \etal~\cite{cao2024security} has demonstrated that adversarial attacks pose substantial real-world threats to critical applications including authentication systems and physiological monitoring services.

%  -> FGSM
The Fast Gradient Sign Method (FGSM)~\cite{goodfellow2014explaining} is a prominent attack technique that generates adversarial perturbations based on gradient signs. Madry \etal~\cite{PGD_Attack} developed the Projected Gradient Descent (PGD) attack method as a variant of FGSM, enabling a broader exploration of the adversarial sample space through iterative updating of FGSM. Carlini-Wagner \etal~\cite{CW} introduced three novel attack methods employing $L_0$, $L_2$, and $L_{\infty}$ distance metrics. Additionally, they proposed the use of high-confidence adversarial adversarial examples as a means to evaluate defense mechanisms.
Croce \etal~\cite{AA} introduced two extensions, namely APGD-CE and APGD-DLR, derived from PGD, to address challenges arising from sub-optimal step sizes and issues associated with the objective function. They further integrated these two attacks with two existing complementary attacks, FAB~\cite{FAB} and Square~\cite{Square}, and then introduced Auto-Attack (AA). AA has become a widely used approach for evaluating model robustness. 

\subsection{Adversarial Training Defense Methods}
Adversarial training (AT) improves the robustness of models by incorporating adversarial examples into the training process, effectively defending against adversarial attacks. 
The conventional AT can be represented as a min-max optimization task, expressed mathematically as:
\begin{equation}
    \underset{\theta}{\min}\mathbb{E}_{(x,y)\sim \mathcal{D}}\underset{||\delta||_p \le \epsilon}{\max}\mathcal{L}(\theta;x+\delta,y),
\end{equation}
where $\mathcal{L}$ represents the loss function with respect to the model's parameter $\theta$, $(x,y)$ is a clean image-label pair sampled from the data distribution $\mathcal{D}$. Additionally, $\delta$ is a perturbation constrained by a maximum $p$-norm magnitude of~$\epsilon$.
\par 
The objective of the inner maximization is to generate adversarial examples, including attacks such as the PGD attack~\cite{PGD_Attack} and other attacks~\cite{attack_1,attack_2,attack_3,attack_4}. 
The outer minimization aims to develop effective training strategies for optimizing network parameters, thereby enhancing the model's robustness against attacks.
AdvProp~\cite{attack_5} enhances robustness by using adversarial examples with separate batch normalization to address distribution differences, improving generalization in image recognition tasks.
Universal adversarial training~\cite{shafahi2020universal} enhances robustness by incorporating universal adversarial perturbations during training, modeling the problem as a two-player min-max game.
Rice \etal~\cite{PGD_AT} conducted a study on the occurrence of overfitting in robust adversarial training and proposed employing a validation set protocol while performing model selection (PGD-AT):
\begin{equation}
    \underset{\theta}{\min}\mathbb{E}_{(x,y)\sim \mathcal{D}}\underset{||\delta||_p \le \epsilon}{\max} \mathcal{L}_{(CE)}(f_{\theta}(x+\delta,y)),
    \label{PGD-AT}
\end{equation}
Zhang \etal~\cite{TRADES} categorized robust errors into “natural errors” and “boundary errors”, balancing robustness and accuracy by controlling the Kullback–Leibler divergence between clean and adversarial outputs~(TRADES):
\begin{multline}
    \underset{\theta}{\min}\mathbb{E}_{(x,y)\sim \mathcal{D}}(\mathcal{L}_{(CE)}(f_{\theta}(x),y)+ \\ \kappa \cdot \underset{||\delta||_p \le \epsilon}{\max}\mathcal{L}_{(KL)}(f_{\theta}(x),f_{\theta}(x+\delta))),
    \label{TRADES_equation}
\end{multline}
Wang \etal~\cite{MART} introduced loss weighting technical based on the classification status to mitigate adversarial misclassification errors while maintaining robustness~(MART):
\begin{multline}
    \underset{\theta}{\min}\mathbb{E}_{(x,y)\sim \mathcal{D}}(\mathcal{L}_{(BCE)}(f_{\theta}(x),y)+ \\ \kappa \cdot \underset{||\delta||_p \le \epsilon}{\max}\mathcal{L}_{(KL)}(f_{\theta}(x),f_{\theta}(x+\delta)) \cdot (1-f_{\theta}(x))),
    \label{MART_equation}
\end{multline}
Wu \etal~\cite{AWP} proposed the flattening adversarial weight loss landscape, enhancing model robustness by directly perturbing network parameters and addressing adversarial vulnerabilities in both input and parameter spaces~(AWP):
\begin{equation}
    \underset{\theta}{\min}\mathbb{E}_{(x,y)\sim \mathcal{D}}\underset{||\delta||_p \le \epsilon, \gamma \in \Gamma}{\max}(\mathcal{L}_{(CE)}(f_{\theta+\gamma}(x+\delta),y)),
    \label{AWP_equation}
\end{equation}
Jia \etal~\cite{jia2022adversarial} introduced  a computationally intensive method that involves automatically generating attacks from a strategy network $g_{\omega}$ and strategy set $\mathcal{A}$ to adaptively find optimal adversarial strategies~(LAS-AT):
\begin{equation}
    \underset{\theta}{\min}\mathbb{E}_{(x,y)\sim \mathcal{D}}\underset{\omega}{\max} \mathbb{E}_{\mathcal{A} \sim p(\mathcal{A}|x;\omega)} \mathcal{L}_{(CE)}(f_{\theta}(x+g_\omega(x,\mathcal{A}),y)),
    \label{LAS_equation}
\end{equation}
These adversarial training approaches exhibit distinct characteristics in their robustness optimization strategies. TRADES and MART represent complementary advances in leveraging clean image properties: TRADES optimizes the KL divergence to balance accuracy and robustness, while MART enhances this framework through adaptive confidence-based weighting for misclassified samples. AWP introduces a distinct paradigm by addressing vulnerability in the parameter space via weight perturbations, moving beyond traditional input-space approaches. LAS-AT further advances the field by introducing a learnable attack strategy network that dynamically adapts adversarial examples during training, representing a significant shift toward adaptive defense mechanisms.
While these approaches have advanced the field through novel attack strategies and training optimizations, they primarily focus on the training process itself, leaving the crucial aspect of feature representation properties during network inference relatively unexplored.

\subsection{Frequency Analysis for Robustness}
Frequency analysis can be utilized to examine adversarial examples and their diverse frequency characteristics during the network inference process. Furthermore, frequency analysis aids in assessing the network's robustness to various frequency features and their influence on the extraction of features. 
Yin \etal~\cite{2019_NIPS_Fourier} demonstrated that data augmentation techniques enhance the robustness of the model against high-frequency distortions while reducing robustness against low-frequency distortions. Furthermore, they suggested that diverse data augmentation methods can alleviate this trade-off.
Wang \etal~\cite{2020_CVPR_Fourier} emphasized that high-frequency components should not be considered mere noise; rather, CNNs can effectively leverage high-frequency information imperceptible to humans. 
Several studies have been influenced by frequency domain analysis and have introduced frequency-based approaches.
Luo \etal~\cite{2022_CVPR_F_work1} imposed low-frequency constraints to restrict adversarial perturbations in high-frequency components, thereby enhancing the stealthiness of attacks. 
Zhou \etal~\cite{2023_CVPR_F_work2} introduced XNet, a semantic segmentation model for biomedical image analysis that incorporates both low-frequency and high-frequency information. 
Bu \etal~\cite{2023_ICCV_frequency} introduced a frequency preference control module that employs the Fourier transform to extract feature maps encompassing low-frequency signals, which facilitates the adjustment of the feature configuration.
However, these methods have disregarded the existence of a low-frequency bias induced by adversarial training. Consequently, they do not effectively utilize and balance the extraction of high-frequency features. This imbalance ultimately results in the irreparable loss of valuable high-frequency information during network inference.

\section{Methodology}
\begin{figure*}[t]
        \begin{center}
        \includegraphics[width=\linewidth]{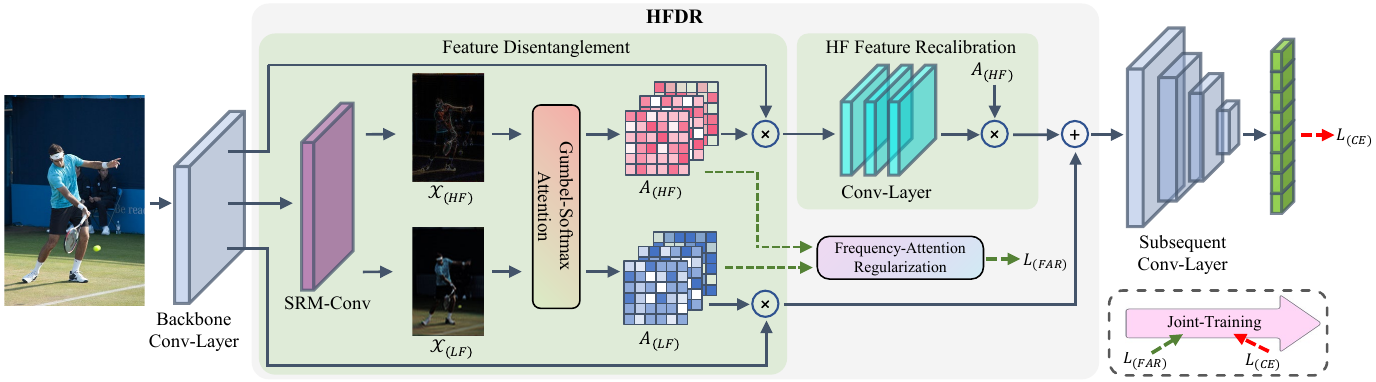}
        \end{center}
        \caption{Overall framework of our proposed method. During network inference, the input feature map is processed through an SRM filter, decomposing it into low-frequency $\mathcal{X}_{(LF)}$ and high-frequency $\mathcal{X}_{(HF)}$ components. To avoid gradient masking, Gumbel-softmax generates element-wise high-frequency $A_{(HF)}$ and low-frequency $A_{(LF)}$ attention maps, facilitating feature disentanglement across frequency domains. After recalibrating high-frequency features, they integrate with low-frequency features before propagation to subsequent network layers. Simultaneously, frequency-aware attention regularization loss $L_{(FAR)}$ constrains the generated attention maps, encouraging the extraction of semantically rich high-frequency features while mitigating low-frequency bias. The model is jointly optimized through $L_{(FAR)}$ and cross-entropy loss $L_{(CE)}$.}
        \label{Model_overview}
\end{figure*}

Adversarial training can introduce an inherent low-frequency bias, limiting the model’s ability to capture meaningful semantic information across different frequency domains. This bias arises from the concentration of adversarial perturbations primarily in the high-frequency domain. To address this issue, we propose a novel High-Frequency Disentanglement and Recalibration (HFDR) module, as illustrated in \Cref{Model_overview}. The HFDR module consists of three key components: feature disentanglement, high-frequency feature recalibration, and frequency-based attention regularization.
\par 
In the feature disentanglement phase (\Cref{Disentanglement}), we generate feature attention maps that selectively focus on distinct frequency domains, enabling the separation of features into high-frequency and low-frequency components. The high-frequency feature recalibration stage (\Cref{Enhance-Com}) then refines the high-frequency information to effectively capture valuable semantic features. Additionally, we introduce a novel regularization technique, Frequency Attention Regularization, to mitigate the low-frequency bias induced by adversarial training (\Cref{attention_regular}).

    \subsection{Feature Disentanglement}
    \label{Disentanglement}

    Feature disentanglement involves separating and isolating distinct components within a feature representation at a specific layer. In network inference, let $\mathcal{X} \in \mathbb{R}^{C \times H \times W}$ denote the feature maps at a predefined layer, where $C$, $H$, and $W$ represent the channel, height, and width dimensions, respectively. Recent approaches commonly rely on low-pass or high-pass filters for feature disentanglement; however, these methods often result in irreversible loss of information~\cite{Filter_1, Filter_2}. To address this limitation, we employ an SRM filter~\cite{SRM} to extract frequency-specific features while ensuring that the proposed disentanglement process retains both high- and low-frequency components:
    \begin{equation}
        \left\{
            \begin{aligned}
            \mathcal{X}_{(HF)} &= \text{Conv}_{1 \times 1}(\text{SRM}(\mathcal{X})), \\
            \mathcal{X}_{(LF)} &= \mathcal{X} - \mathcal{X}_{(HF)},
            \end{aligned}
        \right.
    \end{equation}
    where $\mathcal{X}_{(HF)} \in \mathbb{R}^{C \times H \times W}$ represents the high-frequency feature components aligned across channels through a $1 \times 1$ convolution, and $\mathcal{X}_{(LF)} \in \mathbb{R}^{C \times H \times W}$ represents the low-frequency components calculated as the residual of the original feature maps after subtracting the high-frequency components.
    \par 
    To achieve feature disentanglement while minimizing information loss, we propose using filters to generate element-wise attention maps ${A_i \in \mathbb{R}^{(C \times H \times W)}, i \in \{HF, LF\}}$ for both high-frequency (HF) and low-frequency (LF) semantic sub-features. These attention maps are then used to disentangle the features.
    To address the potential problem that the enhanced model robustness is due to gradient masking~\cite{athalye2018obfuscated}, we incorporate the Gumbel-Softmax method~\cite{jang2016categorical} to generate differentiable attention maps specifically for high-frequency components:
    \begin{equation}
        A_{(HF)} = \frac{\exp{\left(g_{(HF)}+\log(\sigma(\mathcal{X}_{(HF)})/ \tau) \right)}}{\sum_{j\in \{LF, HF\} }  \exp{\left(g_j+\log(\sigma(\mathcal{X}_j))/\tau\right)}},
        \label{Attention}
    \end{equation}
    where $\mathcal{X}_j$ represents distinct semantic sub-features, and $\sigma(\cdot)$ denotes the sigmoid operation. The term $g_j$ incorporates Gumbel noise, defined as $g_j = -\log\left(\log(u_j)\right)$ with $u_j \sim \mathcal{U}(0,1)$. We set $u$ to a fixed value for all samples to control the sampling randomness. The term $\tau$ governs the impact of $g_j$. Additionally, the attention maps for low-frequency features can be computed as $A_{(LF)} = 1 - A_{(HF)}$.
    \par 
    In the context of attention maps $A_{(HF)}$ for high-frequency features, higher attention values indicate a prioritization of capturing semantic information linked to the high-frequency domain, while lower attention values suggest a diminished focus on high-frequency details within features.
    Leveraging frequency-based attention maps allows for the separation of weighted representations of the high-frequency feature $f_{(HF)} \in \mathbb{R}^{(C \times H \times W)}$ and the low-frequency feature $f_{(LF)} \in \mathbb{R}^{(C\times H \times W)}$, defined as:
    \begin{equation}
        \left\{
            \begin{aligned}
            f_{(HF)} &= \mathcal{X} \odot A_{(HF)}, \\
            f_{(LF)} &= \mathcal{X} \odot A_{(LF)},
            \end{aligned}
        \right.
    \end{equation}
    where $\odot$ denotes the Hadamard product.
    
    We acknowledge that adversarial perturbations primarily concentrate in the high-frequency domain. Neglecting these critical inherent cues embedded in high-frequency information may result in missed opportunities for precise predictions. In the following sections, we will investigate the recalibration of high-frequency characteristics to capture valuable cues and semantic information.
  
    \subsection{High-Frequency Feature Recalibration}
    \label{Enhance-Com}
    Previous studies~\cite{high_corruption, long2022frequency, 2019_NIPS_Fourier} have demonstrated that adversarial perturbations mainly target the high-frequency features during network inference. To address this problem, several studies have employed low-pass filtering techniques to mitigate the impact of these high-frequency vulnerabilities and improve model robustness~\cite{low_pass_1, Filter_2}. Nonetheless, it is crucial to acknowledge that these high-frequency characteristics also encompass valuable latent cues for prediction tasks~\cite{high_corruption, high_corruption_1}.
    \par
    Different from previous approaches, our objective is to empower robust models by enabling them to extract features from various frequency domains, rather than simply reducing reliance on vulnerable components. To accomplish this, after disentangling the features into low-frequency and high-frequency components, we suggest recalibrating the high-frequency features to effectively capture the latent semantic clues embedded within them, even in the presence of highly concentrated perturbations. As expressed by the equation below:
    \begin{equation}
        \tilde{f}_{(HF)} = \varphi(f_{(HF)}) \odot A_{(HF)},
    \end{equation}
    where $\varphi$ represents the recalibration network, which is a three-layer convolutional network employed for feature extraction and transformation. The proposed method selectively modifies the activations of high-frequency features $f_{(HF)}$ to mitigate low-frequency biases during the recalibration process. $A_{(HF)}$ serves as a feature attention map to guide the network in the recalibration of high-frequency features, enhancing their discriminative representations. The recalibration network effectively learns and captures informative patterns from the input features through its convolutional layers.
    Following the recalibration of the high-frequency features, we combine the recalibrated high-frequency features $\tilde{f}_{(HF)}$ with the low-frequency features $f_{(LF)}$ to generate the output feature maps  $\tilde{f} = \tilde{f}_{(HF)} + f_{(LF)}$. This fusion process allows for the incorporation of both low and high-frequency information.
    By employing recalibration techniques, we are able to extract valuable predictive cues from high-frequency features that are vulnerable to attacks, while also mitigating potential low-frequency bias introduced by adversarial training.

    % \subsection{Frequency Attention Regularization}
    % \label{attention_regular}
    % Information within the feature map is represented at different frequencies: low-frequency conveys the global structure of image features, while high-frequency unveils the local details and structure. However, previous AT methods~\cite{PGD_AT, AWP, MART, 2023_ICCV_frequency, FAT} often induced frequency biases during network inference. 
    % Without explicit guidance on network frequency preferences, the network may neglect the separated and refined high-frequency features $f_{(HF)}$. 
    % \par
    % Recall that we obtain the high-frequency attention map for each channel $\mathcal{HF}_{(j)} = A_{HF}[j,:,:] \in \mathbb{R}^{(H \times W)}, j \in [1,C]$. We propose a straightforward method termed frequency attention regularization (FAR). The FAR aims to enhance the model's ability to effectively extract both high-frequency and low-frequency features, while simultaneously mitigating inherent low-frequency bias during AT:
    % where $\mathcal{HF}_{(i,j)}$ denotes the attention map for the $j$-th channel of the $i$-th image. $N$ symbolizes the number of images, and $C$ defines the number of channels in the feature maps processed by the HFDR module.
    % Moreover, $\beta$ serves as a balance factor that controls the ratio of information extraction between high-frequency and low-frequency components, while $p$ indicates the norm exponent.

    \begin{algorithm}[t]
    \caption{Adversarial Training with HFDR Module}
    \label{alg:HFDR}
    \begin{algorithmic}[1]
        \REQUIRE Data distribution $\mathcal{D}$, parameters $\theta$, perturbation budget $\epsilon$, trade-off $\lambda$, and balance factor $\varpi$.
        \STATE \textbf{Initialize:} Randomly initialize $\theta$.
        \FOR{each training step}
            \STATE \textbf{Sample data:} $(x, y) \sim \mathcal{D}$.
            \STATE \textbf{Generate adversarial example:} 
            \STATE \hspace{0.35cm} Solve $\delta^\ast = \underset{||\delta||_p \leq \epsilon}{\operatorname{argmax}} \mathcal{L}_{(attack)}(\theta; x + \delta, y)$.
            \STATE \hspace{0.35cm} Compute $x_{\text{adv}} = x + \delta^\ast$.
            \STATE \textbf{Feature extraction:} $\mathcal{X} = \text{Backbone}(x_{\text{adv}})$.
            \STATE \textbf{Feature disentanglement:} 
            \STATE \hspace{0.35cm} Extract low- and high-frequency features $\mathcal{X}_{(HF)}$ and $\mathcal{X}_{(LF)}$ by Eq.~(7).
            \STATE \hspace{0.35cm} Sample frequency-based attention maps $A_{(HF)}$ and $A_{(LF)}$ by Eq.~(8).
            \STATE \hspace{0.35cm} Disentangle frequency features $f_{(HF)} = \mathcal{X} \odot A_{(HF)}$ and $f_{(LF)} = \mathcal{X} \odot A_{(LF)}$.\STATE \textbf{High-Frequency Feature Recalibration:} 
            \STATE \hspace{0.35cm} Obtain $\tilde{f}_{(HF)}$ via capturing latent semantic clue in high-frequency features by Eq.~(10).
            \STATE \textbf{Fusion:} Fuse features: $\tilde{f} = \tilde{f}_{(HF)} + f_{(LF)}$.
            \STATE \textbf{Compute loss:} 
            \STATE \hspace{0.35cm} $\mathcal{L}_{(AT)}$: Adversarial training loss.
            \STATE \hspace{0.35cm} $\mathcal{L}_{(FAR)}$: Promote the model to allocate greater emphasis on high-frequency attention $A_{(HF)}$, alleviating low-frequency bias by Eq.~(11).
            \STATE \hspace{0.35cm} Total: $\mathcal{L} = \mathcal{L}_{(AT)} + \lambda \cdot \mathcal{L}_{(FAR)}$.
            \STATE \textbf{Update:} $\theta \leftarrow \theta - \eta \nabla_\theta \mathcal{L}$.
        \ENDFOR
        \end{algorithmic}
    \end{algorithm}

    \subsection{Frequency Attention Regularization}
    \label{attention_regular}
    
    Feature maps inherently contain multiple frequency components: low-frequency features represent global structures, while high-frequency components capture fine-grained details. In adversarial training, however, prior methods~\cite{PGD_AT, AWP, MART, 2023_ICCV_frequency, FAT} often induce a low-frequency bias, leading the model to favor low-frequency features and overlook high-frequency information. This bias limits the model’s ability to robustly utilize both types of features.
    
    To address this fundamental limitation, we propose the Frequency Attention Regularization (FAR) mechanism, a novel approach that dynamically calibrates the model's attention across the frequency spectrum. FAR introduces a principled framework for maintaining an optimal balance between high- and low-frequency feature extraction, effectively counteracting the inherent low-frequency bias in adversarial training. Our empirical analysis demonstrates that this balanced feature utilization significantly enhances the model's robustness against diverse adversarial attacks.
    
    For each channel, we compute a high-frequency attention map, defined as $\mathcal{HF}{(j)} = A_{HF}[j,:,:] \in \mathbb{R}^{(H \times W)}, j \in [1,C]$. The FAR regularization term is given by:
    
    \begin{equation}
        \mathcal{L}_{(FAR)} = \frac{1}{N \times C}\sum\limits_{i,j}^{N,C}|\frac{||\mathcal{HF}_{(i,j)}||}{||1-\mathcal{HF}_{(i,j)}||}-\varpi |_p,
        \label{loss_item_far}
    \end{equation}
    where $\mathcal{HF}_{(i,j)}$ denotes the high-frequency attention map for the $j$-th channel of the $i$-th image, $N$ represents the batch size, and $C$ indicates the number of channels in the feature maps processed by the HFDR module. The parameter $\varpi$ serves as a critical balance factor that adaptively controls the ratio between high- and low-frequency feature attention, while $p$ determines the norm used in penalizing deviations from the target ratio.
    \par
    The FAR mechanism operates through two key components: (1) the adaptive balance factor $\varpi$, which precisely modulates the emphasis on high-frequency features during network inference, preventing scenarios where $A_{HF}$ fails to adequately capture important high-frequency patterns, and (2) the p-norm penalty, which governs the intensity of correction applied when the ratio between high-frequency and low-frequency features deviates from the target distribution. 

    \subsection{Model Training}
    Our proposed HFDR module can be easily integrated with existing AT methods, such as~\cite{PGD_AT, TRADES, AWP}. By combining the loss function $\mathcal{L}_{(FAR)}$ of the HFDR module with the conventional AT loss function, we derive our objective function as follows:
    \begin{equation}
        \mathcal{L}=\mathcal{L}_{(AT)}+\lambda \cdot \mathcal{L}_{(FAR)},
    \end{equation}
    where $\lambda$ controls the influence of $\mathcal{L}_{(FAR)}$, and $\mathcal{L}_{(AT)}$ represents the loss function associated with various AT models. 
    \par
    By incorporating the HFDR module, we enhance the model's capacity to extract and utilize features across the entire frequency spectrum, leading to improved adversarial robustness. The complete training procedure is detailed in Algorithm \Cref{alg:HFDR}, which provides a step-by-step description of our approach, including the initialization of the HFDR module, the generation of adversarial examples, and the optimization of the combined loss function.
    
\section{Experiments}
\subsection{Experimental Setup}
\input{table/CIFAR_SOTA}

\noindent \textbf{Implementation Details:}
We evaluate the robustness of our approach by conducting experiments on the CIFAR-10~\cite{CIFAR}, CIFAR-100~\cite{CIFAR}, and Imagenette~\cite{Howard_Imagenette_2019} datasets, utilizing the WideResNet34-10~(WRN34-10)~\cite{zagoruyko2016wide}, AlexNet~\cite{wang2024boosting}, and ResNet-18~(RN-18)~\cite{he2016deep} as the baseline models. For the implementation, we insert the HFDR module after the initial convolutional layer. 
To evaluate the impact of HFDR on improving the robustness of various AT techniques,
our model was combined with three AT methods (PGD-AT~\cite{PGD_AT}, MART~\cite{MART}, AWP~\cite{AWP}), denoted as, HFDR-AT, HFDR-MART, and HFDR-AT-AWP. 
We compared our approach with several baseline methods, including, PGD-AT~\cite{PGD_AT}, TRADES~\cite{TRADES}, AWP~\cite{AWP}, MART~\cite{MART}, LAS-AT~\cite{jia2022adversarial}, FSR~\cite{kim2023feature}, FPCM~\cite{2023_ICCV_frequency}, CFA~\cite{CFA} and SGLR~\cite{SGLR}.

We trained WRN34-10 with identical hyperparameters and training details as specified in the original paper~\cite{PGD_AT, MART, AWP}. 
To ensure a fair comparison, we adopted the configuration of PGD-AT~\cite{PGD_AT} as used in the experimental settings of FPCM~\cite{2023_ICCV_frequency}, FSR~\cite{kim2023feature} and CFA~\cite{CFA}.
For ResNet18 and AlexNet architectures, we initialized the learning rate to 0.1. The learning rate was subsequently reduced by a factor of 0.1 at the 90th and 95th epochs. 
The SGD optimizer with a momentum of 0.9 and a weight decay factor of 5e-4 was employed for optimization. 
In the HFDR hyperparameter settings, we set $\lambda=0.1$ and $\varpi=0.1$ for the loss term in \Cref{loss_item_far}. The experiments were conducted on two NVIDIA RTX-A4000 GPUs. 

% Considering the enhanced regularization properties of AWP~\cite{AWP}, we trained AWP-based models with identical hyperparameters and training details as specified in the original paper~\cite{AWP}.

\noindent \textbf{Evaluation Settings:}
We evaluate the robustness of the model using various attack methods, including FGSM~\cite{goodfellow2014explaining}, PGD~\cite{PGD_AT}, CW~\cite{CW}, and AutoAttack~\cite{AA}. Specifically, AutoAttack comprises APGD-DLR~\cite{AA}, APGD-CE~\cite{AA}, FAB~\cite{FAB}, and Square~\cite{Square}. These attacks are performed under the $L_\infty$ norm with $\epsilon=8$.
Notice that, the ``Clean'' denotes the accuracy of clean test samples.

\subsection{Comparison with Other Methods}
\label{robust_test}
In this part, we conducted a comparative analysis of the performance of our proposed module against other methods across diverse dataset conditions, encompassing varying resolutions and dataset sizes. Furthermore, we explored the efficacy of integrating our proposed module with different adversarial training techniques.
\subsubsection{Comparison on CIFAR-10 and CIFAR-100}
HFDR operates as a component that doesn't necessitate intervention during adversarial training and can be integrated with other AT methods to improve robustness. 
\Cref{SOTA_Comparison} presents the efficacy of incorporating the HFDR module in elevating the model's robustness. The key findings are as follows:

(1) HFDR improves the robustness of different adversarial training methods.
After integrating our proposed HFDR module into the three baselines, the HFDR-AT, HFDR-MART, and HFDR-AT-AWP can consistently outperform their corresponding baseline models across all attack scenarios, showcasing superior performance. For example, by incorporating HFDR into the PGD-AT method, we observe performance improvements of 2.60\% in PGD-10, 1.22\% in C\&W, and 2.07\% in AA on CIFAR-10. 
Additionally, there is an increase in clean accuracy by 0.97\% on CIFAR-100. Moreover, integrating the HFDR module into the MART and AWP methods also results in significant improvements in robust performance.
The performance enhancement can be attributed to extracting valuable hidden semantics from high-frequency features that contain concentrated disturbances.

\input{table/Tiny_Imagenet}

(2) Our model outperforms the baseline under attack.
Our proposed model, HFDR-AT-AWP, exhibits superior performance compared to all baseline methods, showcasing exceptional robustness. Specifically, HFDR-AT-AWP outperforms the previous best baseline AWP by 1.77\% of PGD-10, 1.51\% of PGD-50, and 2.34\% of AA on CIFAR-100. Besides, our method significantly outperforms the state-of-the-art frequency-based method, FPCM by a large margin. Unlike FPCM, which inserts multiple blocks into the network, we only inserted a single block, reducing the additional computation cost. 
       
(3) HFDR impacts the generalization ability on clean samples.
% Our model demonstrates substantial advancements in robustness against a wide range of attacks. However, it is important to acknowledge that 
HFDR may introduce instability in the model's generalization performance on clean samples.
Specifically, we note that HFDR-AT improved clean accuracy on CIFAR-10 and CIFAR-100, with gains of 0.72\% and 0.97\%, respectively. Conversely, the HFDR-MART approach exhibits a decrease of 0.74\% in clean sample accuracy on CIFAR-10, combined with a modest increase of 0.93\% on CIFAR-100.
One plausible explanation is that finer-grained and high-resolution data can help HFDR generate more distinctive attention maps, enabling effective recalibration.  This ability allows the HFDR module to generate discriminative attention maps during network inference, enhancing the model's capability to extract valuable cues from vulnerable high-frequency features.
% One plausible explanation for this observation is the finer granularity of labels in CIFAR-100. This characteristic enables the HFDR module to generate more discriminative attention maps during network inference. 
% As a consequence, the model employing the HFDR module is efficiently configured to extract latent and useful cues within vulnerable high-frequency features.

\input{table/Transfer_attack}
\subsubsection{Comparison on Tiny ImageNet}
To evaluate the credibility and generalizability of the performance improvements, the corresponding performance results on the Imagenette~\cite{deng2009imagenet} are presented in \Cref{ImageNet-Robust}. This dataset features higher resolution compared to CIFAR-10 and CIFAR-100. Specifically, the evaluation results on the Imagenette dataset reveal that HFDR-AT achieved performance enhancements of 2.75 and 2.86\% against C\&W and AA attacks, respectively, compared to PGD-AT. Furthermore, HFDR-AT-AWP exhibited improvements of 1.99\% and 3.61\% when compared to AWP. These findings suggest that our approach seamlessly integrates with adversarial training frameworks, maintaining robust performance when confronted with complicate datasets.

\subsection{Robustness to Transfer Attacks}
In this part, we evaluated the performance of models equipped with the HFDR module against transfer attacks. When adversaries lack access to network parameters, they may employ alternative source models to craft adversarial examples for targeting the model under consideration. By scrutinizing the model's robustness against transfer attacks, we aim to demonstrate that the enhancements in performance are not  attributable to gradient masking.
We conducted evaluations on the AlexNet and WRN34-10 models equipped with the HFDR module, utilizing techniques involving ResNet50, VGG16, Inception-v3 models, as well as PGD-10, C\&W, AutoAttack (AA), and Spectrum Simulation Attack (SSA) to generate adversarial examples. The results of these assessments are presented in \Cref{trans_attack}, leading to the following conclusions:

(1) HFDR improve the robustness against various transfer attack.
Models enhanced with the HFDR module have exhibited improved performance when encountering various transfer attacks generated by different source models. For instance, in scenarios involving source models ResNet50 and Inc-v3, the AlexNet equipped with HFDR showed a performance increase of 1.52\% and 0.70\% compared to the baseline model when exposed to AutoAttack (AA), while the WRN34-10 equipped with HFDR demonstrated performance gains of 1.06\% and 2.18\% when subjected to CW attacks. This observation further highlights that the performance enhancement achieved through our method is not attributed to gradient masking.

(2) HFDR improve the performance against the frequency-based attack.
Furthermore, we evaluated the models' robustness to frequency-domain attack SSA~\cite{long2022frequency}. SSA employs discrete Fourier transform and inverse discrete Fourier transform to produce transferable attack instances. Specifically, when the AlexNet model, enhanced with the HFDR module, faced SSA attacks originating from ResNet50 and Inc-v3, the defense success rates improved by 1.46\% and 1.04\% respectively compared to the baseline approach.

\subsection{Ablation Analysis}

\input{table/Ablation}

\subsubsection{Evaluation on HFDR module}
We evaluated the performance of using the disentanglement and recalibration process of high-frequency features (HFDR-Net) and adding the frequency attention regularization term (HFDR-FAR) in our proposed HFDR.
From \Cref{tb:Ablation}, we can find that: 

(1) HFDR-Net capture latent high-frequency semantics.
We observed that including the HFDR-Net network can lead to significant improvements in the standard accuracy and robustness of our model. This discovery not only confirms the potential of capturing predictive semantic information from high-frequency features but also indirectly validates the limitations of traditional adversarial training methods in extracting high-frequency features. 

(2) HFDR-Regularization mitigate low-frequency bias.
Incorporating frequency domain attention-based regularization has led to enhancements across various metrics. For instance, compared to the scenario without this loss term, the performance of PGD-10 attack and AA attack has shown respective increases of 0.95\% and 0.60\%. This highlights the efficacy of frequency domain attention regularization in mitigating low-frequency biases during adversarial training.

\input{table/Ablation_feature}
\input{table/DFT_table}

\subsubsection{Effectiveness of different frequency components}
We analyzed the role played by the different frequency components of HFDR in the disentangling and recalibrating process during network inference. 
From \Cref{tb:frequency_components}, we can find that:
    
(1) Leveraging only high-frequency features can yield high accuracy.
The experimental results demonstrate that a classification accuracy of 76.56\% can be achieved, even when utilizing solely uncalibrated high-frequency features $h_{(HF)}$. These features also exhibited robustness against PGD-10 (51.12\%) and AA (46.27\%). 
The findings of this study provide additional confirmation of the viability of our work, demonstrating that the network possesses the ability to accurately capture semantically significant information for prediction from high-frequency features, even when subjected to highly concentrated perturbations. 

(2) Recalibration effectively captures valuable high-frequency features.
Compared to $f_{(HF)}$, training with the recalibrated high-frequency features $\tilde{f}_{(HF)}$ yielded improvements of 2.41\% and 1.09\% under the FGSM and PGD-10 attacks, respectively. This demonstrates that our recalibration stage effectively captures valuable semantics that enhance the predictive capabilities of the model.

(3) HFDR harmonizes the extraction of high-frequency and low-frequency features.
By employing HFDR module to obtain $\tilde{f}=\tilde{f}_{(HF)}+f_{(LF)}$, we observed significant improvements in multiple performance aspects compared to the initial features $f=f_{(HF)}+f_{(LF)}$. For example, $\tilde{f}$ exhibits enhancements of 2.77\% and 1.23\% under PGD-10 and AA attacks, respectively.

\subsubsection{Effectiveness of different frequency filtering}
The effectiveness of feature disentanglement using various frequency domain filtering methods is examined in \Cref{Frequency-Techncal}. Specifically, the DFT-based approach involves processing data transformed by the Discrete Fourier Transform, where the spectral central width is denoted as B. In this procedure, frequency components located outside of a spectral central square are zeroed out, followed by the utilization of the inverse Discrete Fourier Transform for filtering. The findings suggest that high-pass filters with a smaller central width B exhibit a more improved performance. This can be attributed to its enhanced capability to separate high-frequency features from image features. 
By focusing on the high-frequency components that capture intricate and distinctive details, the filter DFT(B=8) augment the model's ability to extract subtle patterns and variations in the data during feature disentanglement, leading to an improved performance in feature recalibration. Nevertheless, these results do not surpass the performance of the SRM filter.

\input{table/Position}
\begin{figure}[t]
    \begin{center}
    \includegraphics[width=\linewidth]{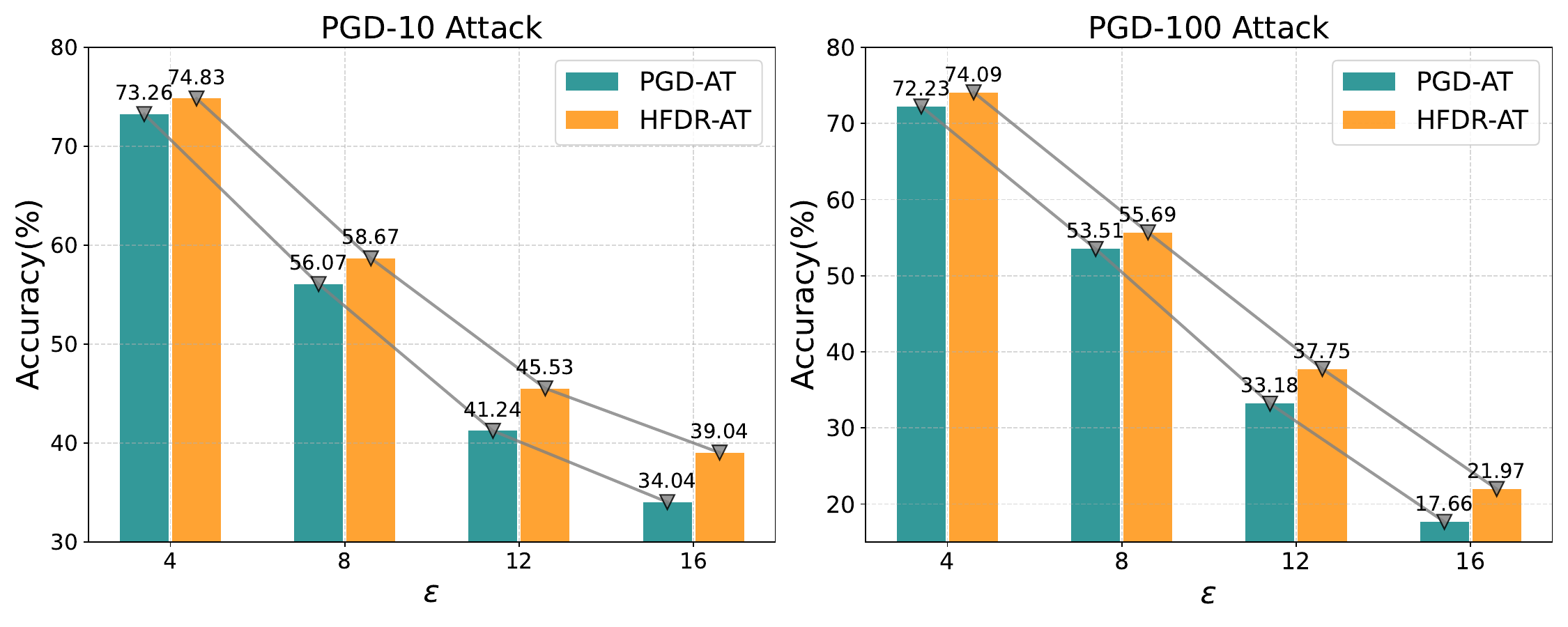}
    \end{center}
    \caption{Comparison of robust accuracy using WRN34-10 on CIFAR-10 dataset under PGD-10 and PGD-100 attacks with varying $\epsilon$ values. The $x$-axis shows the perturbation bound $\epsilon$, and the $y$-axis shows the robust accuracy (\%).}
    \label{Eplison_ana}
\end{figure}

\begin{figure*}[th]
        % \begin{center}
    \centering
    \includegraphics[width=0.8\linewidth]{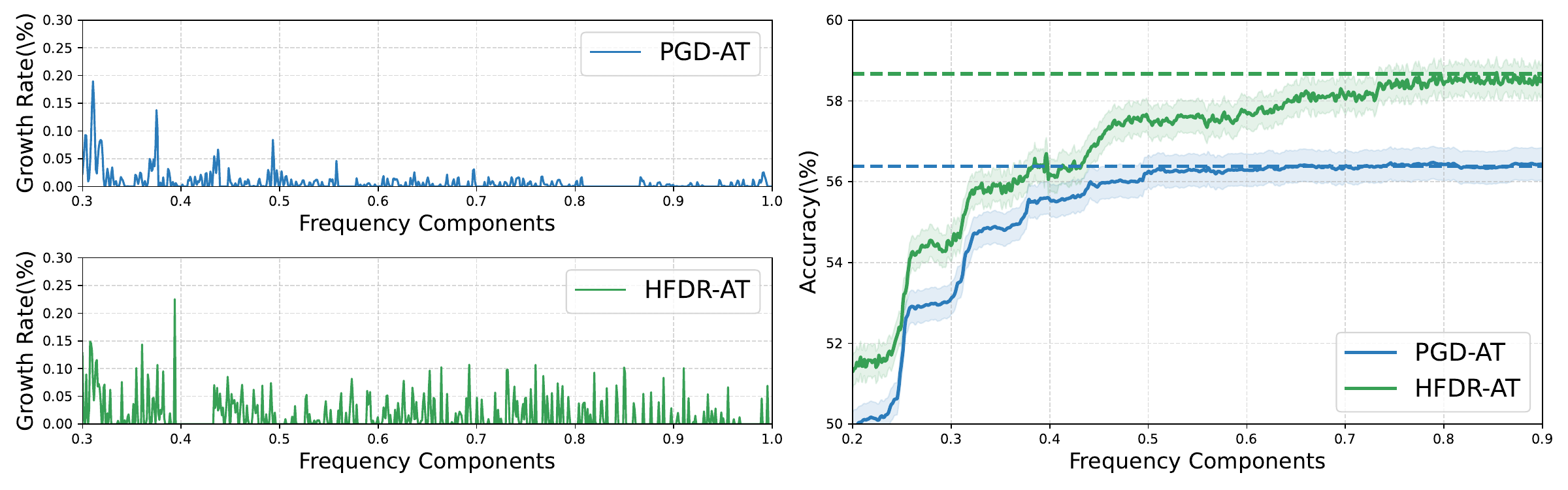}
        % \end{center}
        \caption{Comparison of growth rate (\%) and robust accuracy (\%) under PGD-10 attack ($\epsilon=8$) between HFDR-AT and PGD-AT methods across increasing frequency components.}
        \label{Model_robust_freq_com}
\end{figure*}

\begin{figure}[t]
    \begin{center}
    \includegraphics[width=\linewidth]{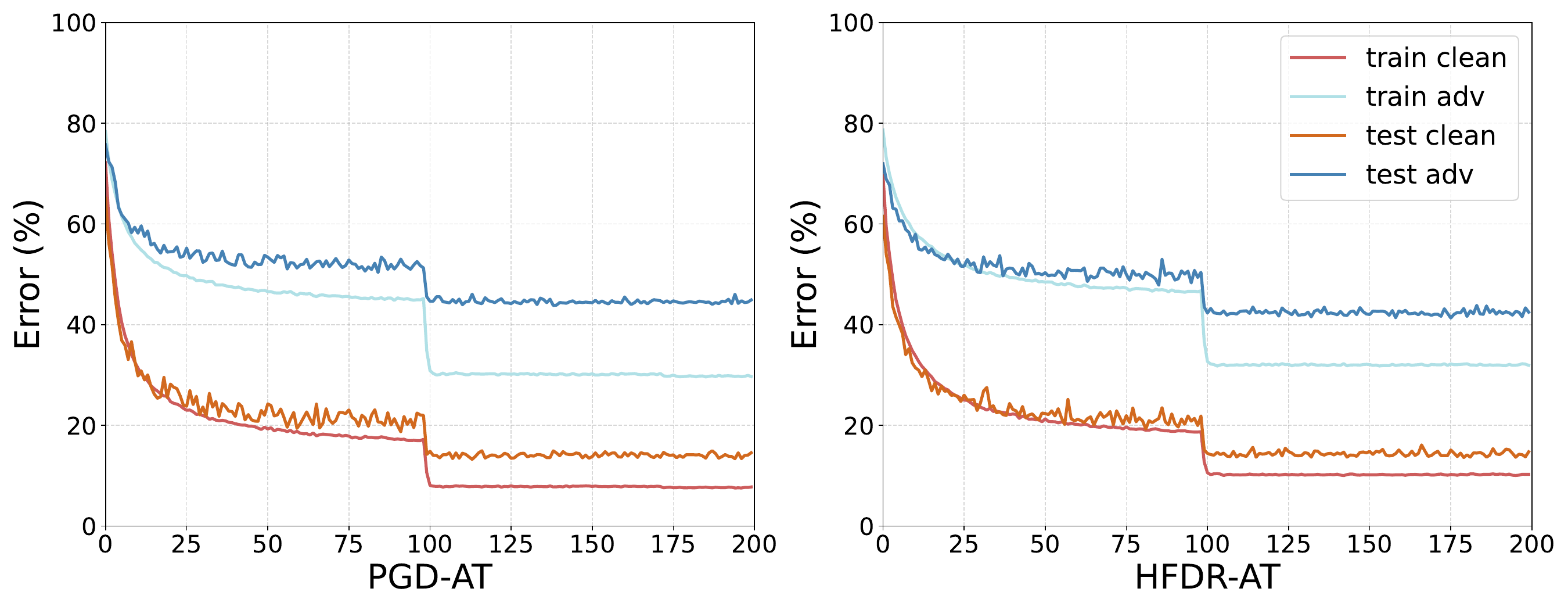}
    \end{center}
    \caption{Learning curves showing clean and robust accuracy on both training and test sets of CIFAR-10 dataset using WRN34-10 architecture.}
    \label{Robust_overfitting_figure}
\end{figure}

\subsubsection{Robustness at Different HFDR Positions}
The robustness of WRN34-10  with diverse HFDR module placements is presented in \Cref{Layer-Robust}. The outcomes reveal that integrating the HFDR module after Conv.1 leads to exceptional robust and clean performance. This effect is attributed to the relationship between the network's depth and the level of abstraction in the extracted features. Features with shallower depths and lower levels of abstraction are better suited for discerning different frequency characteristics during feature decoupling, allowing the HFDR module to perform more effective high-frequency feature calibration.

\subsection{Robust Generalization Analysis}
\subsubsection{Robustness under Various $\epsilon$ Attacks}
We investigated the robustness of our proposed HFDR module under different levels of attack intensity. From the results in~\cref{Eplison_ana}, we can have the following observations: 

(1) HFDR enhances robustness against adversarial attacks. 
Our proposed method demonstrates improved performance against FGSM and PGD-10 attacks across different values of $\epsilon$. For example, under the PGD-10 attack, HFDR-AT shows respective increases of 1.57\%, 2.60\%, 4.29\%, and 5.00\% compared to PGD-AT at $\epsilon$ values of 4, 8, 12, and 16.

(2) HFDR enhances the generalization against attacks.
As the value of $\epsilon$ increases, our method demonstrates improved generalization capabilities by exhibiting a smaller decrease in accuracy compared to the baseline approach. For instance, when $\epsilon$ increases from 8 to 16, PGD-AT experiences a 22.03\% decrease in accuracy, whereas HFDR-AT only observes a 19.63\% decline under the PGD-10 attack.

\subsubsection{Robust Generalization Gap}
Our investigation delved into the influence of HFDR on the deviation in robust generalization during the training process. We extended the training epochs to 200 to enhance the visualization effect, as illustrated in \cref{Robust_overfitting_figure}. It is evident that HFDR-AT significantly alleviates overfitting in the model during training in comparison to PGD-AT. Furthermore, it diminishes both the discrepancy in robust generalization gap and standard generalization gap, thereby enhancing the model's generalization capability.

\begin{figure}[th]
    \begin{center}
    \includegraphics[width=\linewidth]{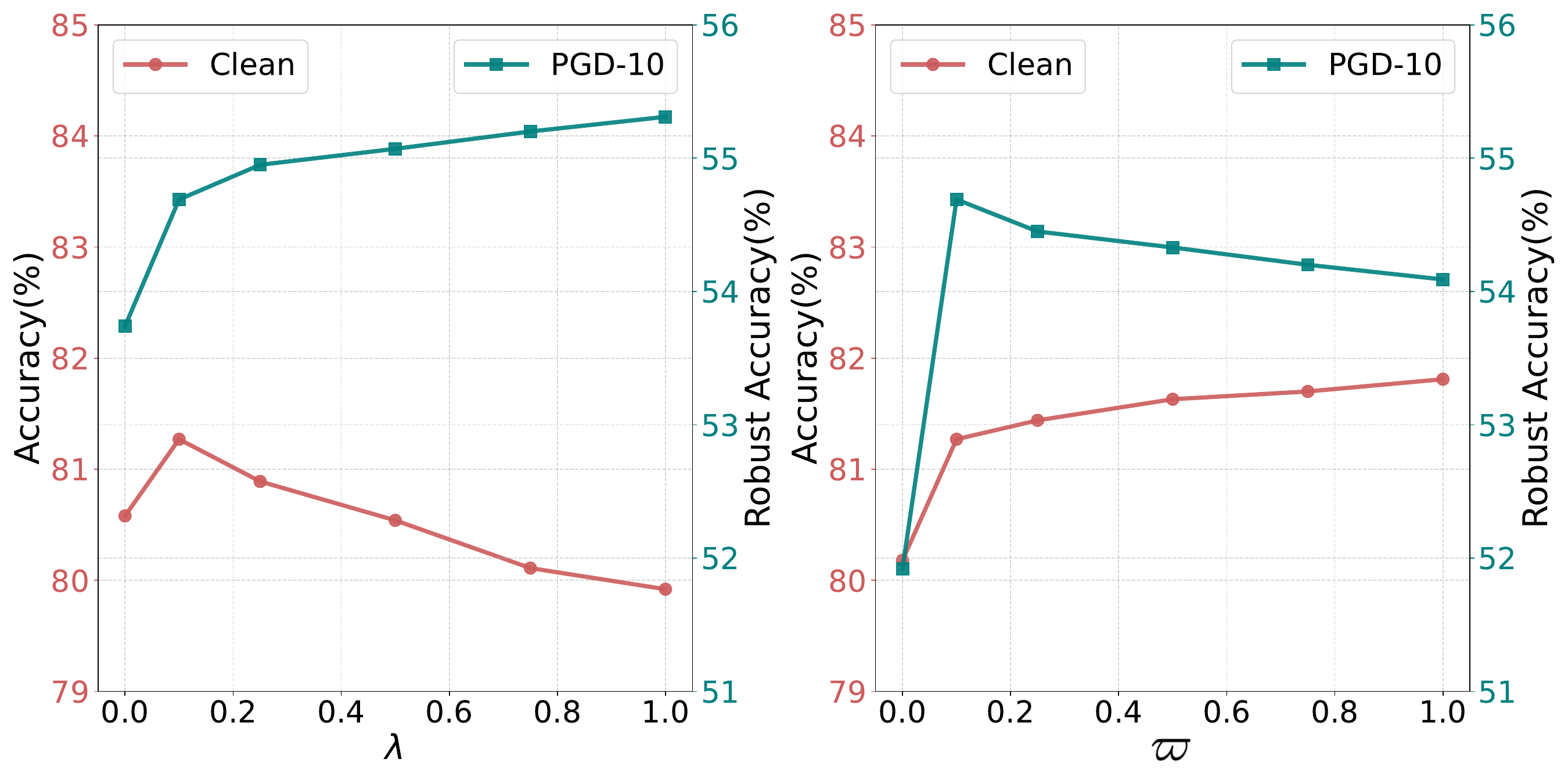}
    \end{center}
    \centering
    \caption{Analysis of hyper-parameters $\lambda$ and $\varpi$ using ResNet-18 on CIFAR-10. The $x$-axis represents the parameter value, while the $y$-axis represents the accuracy(\%).} 
    \label{Parameter_ana}
\end{figure}

\begin{figure}[th]
    \begin{center}
    \includegraphics[width=\linewidth]{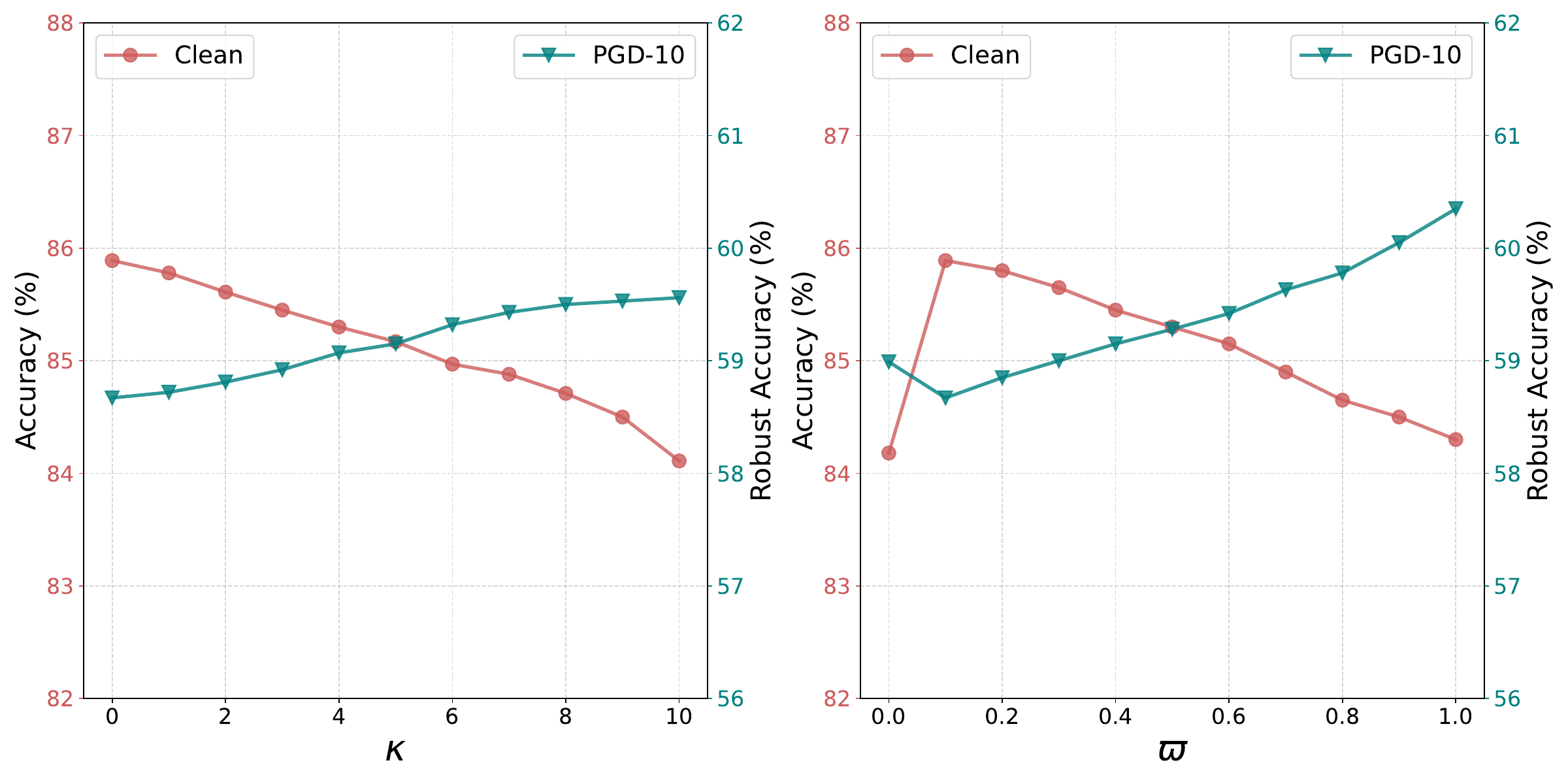}
    \end{center}
    \centering
    \caption{Analysis of hyper-parameters $\kappa$ and $\varpi$ using ResNet-18 on CIFAR-10. The $x$-axis represents the parameter value, while the $y$-axis represents the accuracy(\%).} 
    \label{Parameter_ana1}
\end{figure}

\subsection{Parameter Sensitive Analysis}
In this part, we performed an analysis of the HFDR module hyper-parameters. In \Cref{Parameter_ana}, a visualization of parameter sensitivity for hyperparameters $\lambda$ and $\varpi$ in the HFDR method is demonstrated. 
Additionally, we also analyzed the impact of $\kappa$ and $\varpi$ (HFDR-MART) in \Cref{Parameter_ana1}.
Notably, assigning a value of zero to $\varpi$ results in the absence of the frequency attention regularization term (refer \Cref{loss_item_far}). Evidence suggests that these two hyper-parameters influence the trade-off between model clean and robust accuracy performance. The following offers a detailed analysis:

(1) The parameter $\lambda$ determines the magnitude of the frequency attention regularization term. 
As the parameter $\lambda$ increases, there is a corresponding rise in robust accuracy, contrasted by a decline in clean accuracy. When the value of $\varpi$ is maintained constant, an increase in $\lambda$ causes the the adversarial training focus on high-frequency separation, , which can slightly compromising the model's capability to extract features from clean examples.
However, a larger $\lambda$ value facilitates the disentanglement of high-frequency components during network inference, thereby enhancing the model’s capability to recalibrate perturbed high-frequency features. To strike an optimal balance between model robustness and clean accuracy, it is recommended to set $\lambda=0.1$.

(2) The parameter $\varpi$ influences the extraction of high-frequency features. Increasing $\varpi$ typically reduces robust accuracy while enhancing clean accuracy. Notably, the model performs better when the frequency attention regularization term is activate (\ie  $\varpi \neq 0$).
As $\varpi$ increases, $f_{(HF)}$ retains more high-frequency feature information, improving the model’s ability to recalibrate these informative features in clean images.
Conversely, adversarial attacks introduce perturbations that disrupt high-frequency features, complicating their recalibration. Therefore, a balanced setting, such as $\varpi = 0.1$, is recommended.

(3) The parameter $\kappa$ controls the trade-off between clean accuracy and robust accuracy. In the left figure, we fixed $\varpi=0.1$ and analyzed how varying $\kappa$ affects both clean accuracy and robustness. The right figure assesses the impact of varying $\varpi$ while keeping $\kappa=6$. For these two figures, we find that increasing $\kappa$ or $\varpi$ will generally decrease clean accuracy while increasing robustness. To have a balance, we choose $\kappa=6$ (following~\cite{2023_ICCV_frequency,kim2023feature,jia2022adversarial}) and $\varpi=0.1$ in this study to have a balance between clean accuracy and robust accuracy.

\input{table/stable}

\begin{figure}[t]
        \begin{center}
        \includegraphics[width=\linewidth]{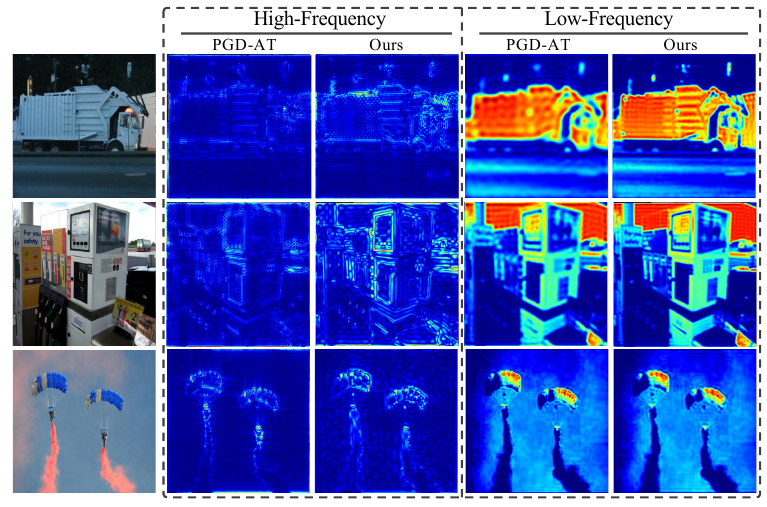}
        \end{center}
        \caption{Visualization of the ability of HFDR-AT (ours) and traditional adversarial training (PGD-AT) to extract features across different frequency components. The results demonstrate the feature map activations before the Conv2 layer of ResNet-18 on the Imagenette dataset.}
        \label{Fre_attention}
\end{figure}

\subsection{Analysis of Mitigation of Low-Frequency Bias}
In this part, we delve into exploring the mitigating effects of our proposed module HFDR on low-frequency biases during adversarial training. In \Cref{Model_robust_freq_com}, we showcased the efficacy of our HFDR-AT method in extracting diverse frequency domain features compared to PGD-AT. The left figure illustrates that the growth trend in model performance with the incorporation of low to high-frequency features, whereas the right figure depicts variations in model performance with the incorporation of low to high-frequency components. 
Additionally, as visualized in \Cref{Fre_attention}, our method significantly enhances the extraction of high-frequency features while maintaining the ability to capture low-frequency features. These results indicate that our model recalibrates vulnerable high-frequency features, mitigating the low-frequency bias inherent in traditional adversarial training.

\subsection{Evaluation of Experimental Stability}
To evaluate the stability of our experimental results, we performed five independent runs of each experiment and reported the mean and standard deviation of the outcomes. As shown in \Cref{stable}, the results exhibit minimal fluctuations across multiple runs, confirming the robustness and consistency of the HFDR method under varying conditions.

\subsection{Computational Efficiency}
\Cref{tb:computation} depicts the comparison of our approach and the original model in terms of training efficiency. This comparison includes the analysis of the model parameters and the average time required for one training cycle. The results highlight that our model enhances the robustness of the model without a significant increase in model parameters or the cost of AT. For instance, on the CIFAR-10, the regular WRN34-10 takes 1,050 seconds for one training cycle, whereas the model with HFDR only takes 1,068 seconds. This demonstrates the clear advantages of our model compared to other methods with complex network architectures.

\input{table/Computation_cost}

\section{Conclusion}
    In this paper, we presented a comprehensive analysis of frequency-domain characteristics in adversarially trained networks and introduced HFDR (High-Frequency Disentanglement and Recalibration), a novel approach to enhance model robustness. Our investigation revealed a critical low-frequency bias in conventional adversarial training methods, where models predominantly rely on low-frequency components while overlooking valuable semantic information contained in high-frequency features.
    To address this limitation, we developed the HFDR module, which employs high-pass filters and frequency attention mechanisms to effectively disentangle and recalibrate high-frequency features. Through the integration of frequency domain attention regularization, our approach successfully harmonizes feature extraction across different frequency domains, mitigating the inherent low-frequency bias of adversarial training. Extensive experiments verified the effectiveness of our method.
    \par
    On the other aspect, this paper could provide a notable guiding direction for future research on model robustness. We introduce a novel perspective that emphasizes the importance of the robustness of feature extraction capability in achieving outstanding model robustness. In other words, when confronted with perturbations across various frequency domain features, a robust model should consistently extract valuable semantic information from these features rather than neglecting the robustness provided by features with high-density perturbation frequencies. 

\bibliographystyle{IEEEtran}
\bibliography{reference}

\end{document}

%% file: table/CIFAR_SOTA.tex
\begin{table*}[t]
        \begin{center}
        \caption{Test robustness (\%) using WRN34-10. The \textbf{number} in bold indicates the best accuracy.}
        \label{SOTA_Comparison}
        \resizebox{0.95\textwidth}{!}
        {
            \begin{tabular}{clccccccc}
            \toprule
            Dataset&Method& Publish& Clean& PGD-10& PGD-20& PGD-50& C\&W& AA\\ \midrule
            \multicolumn{1}{c}{\multirow{12}{*}{CIFAR-10}}&PGD-AT~\cite{PGD_AT}             &ICML-20& 85.17& 56.07& 55.08& 54.88& 53.91& 51.69\\ 
            &TRADES~\cite{TRADES}             &ICML-19& 85.72& 56.75& 56.10& 55.90& 53.87& 53.40\\ 
            &AWP~\cite{AWP}                   &NeurIPS-20& 85.57& 58.92& 58.13& 57.92& 56.03& 53.90\\
            &MART~\cite{MART}                 &ICLR-20& 84.17& 58.98& 58.56& 58.06& 54.58& 51.10\\ 
            &LAS-AT~\cite{jia2022adversarial} &CVPR-22& \textbf{86.23} &57.64 & 56.49& 56.12& 55.73& 53.58 \\
            &FSR~\cite{kim2023feature}        & CVPR-23 & 84.46 & 57.17 & 56.70 & 56.51 & 54.76 & 53.03\\
            &FPCM~\cite{2023_ICCV_frequency}  &ICCV-23  & 85.46 & 56.93 & 56.44 & 56.18 & 54.40 & 52.86\\ 
            &CFA~\cite{CFA}                   & CVPR-23 & 84.90 & 57.78 & 57.35 & 56.81 & 54.56 & 52.63 \\
            &SGLR~\cite{SGLR}                 & CVPR-24 & 85.81 & 57.83 & 57.40 & 56.99 & 54.63& 52.70\\
             \cmidrule{2-9}
            &HFDR-AT                          &\multicolumn{1}{c}{\multirow{4}{*}{Ours}}&85.89& 58.67& 57.69& 56.94& 55.13& 53.76\\
            &HFDR-TRADES                          &&84.18&58.99&58.40& 58.31& 55.43& 53.86\\
            &HFDR-MART                            &&83.43& 59.73& 58.96& 58.73&   54.56 & 53.92\\
            &HFDR-AT-AWP                          && 85.19& \textbf{60.46}& \textbf{59.28}& \textbf{58.76}& \textbf{56.62}& \textbf{54.64}
            \\ \midrule
            \multicolumn{1}{c}{\multirow{12}{*}{CIFAR-100}}&PGD-AT~\cite{PGD_AT}   &ICML-20& 60.89& 32.19& 31.69& 31.45& 30.10& 27.86\\ 
            &TRADES~\cite{TRADES}   &ICML-19& 58.61& 29.20& 28.66& 28.56& 27.05& 25.94\\
            &AWP~\cite{AWP}         &NeurIPS-20& 60.38& 34.13& 33.86& 33.65& 31.12& 28.86\\ 
            &MART~\cite{MART}       &ICLR-20& 59.23& 33.10& 32.77& 32.56& 30.32& 28.57\\ 
            &LAS-AT~\cite{jia2022adversarial} 
                                    &CVPR-22& 61.80& 33.45& 32.77& 32.54& 31.12& 29.03\\
            &FSR~\cite{kim2023feature}        
                                    & CVPR-23 & 59.41& 33.76& 33.33& 32.94& 30.40& 28.56\\
            &FPCM~\cite{2023_ICCV_frequency}  
                                    &ICCV-23& 60.97& 32.57& 32.18& 32.90& 30.27& 28.15\\ 
            &CFA~\cite{CFA}         &CVPR-23& 61.26& 33.18& 32.67& 32.55& 30.74& 29.40\\
            &SGLR~\cite{SGLR}       &CVPR-24& 61.10& 33.80& 33.52& 33.27& 30.88& 29.51\\
            \cmidrule{2-9}
            &HFDR-AT                &\multicolumn{1}{c}{\multirow{4}{*}{Ours}}& \textbf{61.86}& 35.23& 34.61& 34.29& 31.13& 30.27\\
            &HFDR-TRADES            &&60.27&35.57&34.93&34.53&31.50&30.59\\
            &HFDR-MART              &&60.16& 35.94& 35.60& 35.41& 32.07& 30.98\\
            &HFDR-AT-AWP            &&60.97& \textbf{36.95}& \textbf{36.47}& \textbf{36.16}& \textbf{32.70}& \textbf{31.54}  
            \\ \bottomrule
            \end{tabular}
        }
        \end{center}
    \end{table*}

%% file: table/Tiny_Imagenet.tex
\begin{table}[t]
	\begin{center}
		\caption{Accuracy(\%)  on Imagenette using ResNet18. The \textbf{number} in bold indicates the best accuracy.}
		\label{ImageNet-Robust}
		\resizebox{\linewidth}{!}
		{
			\begin{tabular}{l|cccccc}
				\hline
				Method           & Clean          & PGD-10         & PGD-20         & PGD-50         & C\&W           & AA             \\ \hline 
				PGD-AT~\cite{PGD_AT}        &83.80&58.11&57.56&57.08&55.35&54.69 \\
                    MART~\cite{MART} &81.17&59.71&59.23&59.03&57.17&56.18\\
                    AWP~\cite{AWP}  &83.25&61.32&60.92&60.74&59.39&57.05\\
				\hline
				HFDR-AT    &85.39&60.59&60.24&59.46&58.10&57.55\\ 
                    HFDR-MART  &\textbf{86.22}&62.50&62.18&61.84&60.71&59.56\\
                    HFDR-AT-AWP  &84.95&\textbf{62.94}&\textbf{62.77}&\textbf{62.41}&\textbf{61.38}&\textbf{60.67}\\
				\hline
			\end{tabular}
		}
	\end{center}
\end{table}

%% file: table/Transfer_attack.tex
\begin{table*}[htp]
    \caption{Transfer attack accuracy~(\%) in the single-model transfer scenario. The \textbf{number} in bold indicates the best accuracy.}
    \label{trans_attack}
    \begin{center}
    \resizebox{0.9\textwidth}{!}
    {
    \begin{tabular}{c|ccc|ccc}
        \toprule
        \multirow{3}*{Attack ($\epsilon=8$)} & \multicolumn{6}{c}{Performance w/o and w/ HFDR}\\
        \cmidrule{2-7}
         & \multicolumn{3}{c|}{Source: AlexNet} & \multicolumn{3}{c}{Source: WRN34-10} \\
        & $\Rightarrow$ ResNet50 &$\Rightarrow$ VGG16 & $\Rightarrow$ Inc-v3&$ \Rightarrow$ ResNet50 &$\Rightarrow$ VGG16 & $\Rightarrow$ Inc-v3\\ \midrule
        PGD-10  &59.45/\textbf{60.50}&48.14/\textbf{49.47}&50.91/\textbf{51.88}&65.05/\textbf{65.92}&69.51/\textbf{70.44}&65.48/\textbf{65.99}\\
        C\&W      &59.49/\textbf{60.47}&48.29/\textbf{48.77}&51.68/\textbf{52.64}&64.70/\textbf{64.88}&68.76/\textbf{70.24}&64.39/\textbf{65.03}\\
        AA      &60.82/\textbf{62.34}&53.45/\textbf{54.01}&56.31/\textbf{57.01}&70.47/\textbf{71.53}&66.97/\textbf{68.85}&67.16/\textbf{69.34}\\
        SSA             &35.30/\textbf{36.76}&32.88/\textbf{33.62}&37.57/\textbf{38.61}&44.63/\textbf{45.91}&48.91/\textbf{50.35}&43.84/\textbf{44.58}\\
        \bottomrule
    \end{tabular}
    }
    \end{center}
\end{table*}

%% file: table/Ablation.tex
\begin{table}[t]
        \begin{center}
        \caption{Comparison of accuracy (\%) with and without different HFDR components against various adversarial attacks. The \textbf{number} in bold indicates the best accuracy.}
        \vspace{-0em}
        \label{tb:Ablation}
        \resizebox{\linewidth}{!}
        {
            \begin{tabular}{lcccccc}
            \toprule
            \textbf{Source:  RN-18} & \multicolumn{1}{c}{\multirow{2}{*}{Clean}} &\multicolumn{5}{c}{Attack}\\
            \cmidrule{3-7}
            Method & & FGSM&PGD-10&PGD-50&C\&W&AA\\ 
            \midrule
            Original&80.16&74.58&51.92&50.95&48.90&47.10 \\
            +HFDR-Net     &80.58&75.73&53.74&52.95&50.99&47.83 \\
            \rowcolor{gray!20}
            +HFDR-FAR &\textbf{81.27}&\textbf{76.64}&\textbf{54.69}&\textbf{53.58}&\textbf{51.86}&\textbf{48.33} \\
            \bottomrule
            \end{tabular}
        }
        \end{center}
        \vspace{-0em}
    \end{table}

%% file: table/Ablation_feature.tex
\begin{table}[t]
        \begin{center}
        \caption{Ablation analysis on the feature extract capability of various features obtained throughout HFDR-AT. ``Method'' represents the feature components for network inference during training. The \textbf{number} in bold indicates the best accuracy.}
        \vspace{-0em}
        \label{tb:frequency_components}
        \resizebox{\linewidth}{!}
        {
            \begin{tabular}{lcccccc}
            \toprule
            \textbf{Source:  RN-18} & \multicolumn{1}{c}{\multirow{2}{*}{Clean}} &\multicolumn{5}{c}{Attack}\\
            \cmidrule{3-7}
            Method & & FGSM&PGD-10&PGD-50&C\&W&AA\\ 
            \midrule
            $f$&80.16&74.58&51.92&50.95&48.90&47.10\\
            $f_{(HF)}$&76.54&72.02&51.12&50.03&47.76&46.27\\
            $f_{(LF)}$&78.54&73.14&51.41&51.67&47.86&46.13\\
            $\tilde{f}_{(HF)}$&79.67&74.43&52.21&51.40&48.14&46.82 \\
            \rowcolor{gray!20}$\tilde{f}$(ours)&\textbf{81.27}&\textbf{76.64}&\textbf{54.69}&\textbf{53.58}&\textbf{51.86}&\textbf{48.33} \\
            \bottomrule
            \end{tabular}
        }
        \end{center}
        \vspace{-0em}
    \end{table}

%% file: table/DFT_table.tex
\begin{table}[t]
        \begin{center}
        \caption{The effect of HFDR with DFT high-pass filter. The \textbf{number} in bold indicates the best accuracy.}
        \vspace{-0em}
        \label{Frequency-Techncal}
        \resizebox{\linewidth}{!}
        {
            \begin{tabular}{lcccccc}
            \toprule
            \textbf{Source:  RN-18} & \multicolumn{1}{c}{\multirow{2}{*}{Clean}} &\multicolumn{5}{c}{Attack}\\
            \cmidrule{3-7}
            Method & & FGSM&PGD-10&PGD-50&C\&W&AA\\ 
            \midrule
            N/A &80.16&74.58&51.92&50.95&48.90&47.10\\ 
            DFT(B=8)&81.11&75.99&55.25&52.16&50.45&47.81\\
            DFT(B=16)&81.12&76.07&54.02&51.21&49.97&47.60\\ 
            \rowcolor{gray!20}
            SRM(Ours)&\textbf{81.27}&\textbf{76.64}&\textbf{54.69}&\textbf{53.38}&\textbf{51.86}&\textbf{48.33}\\ 
            \bottomrule
            \end{tabular}
        }
        \end{center}
        \vspace{-0em}
    \end{table}

%% file: table/Position.tex
\begin{table}[t]
        \begin{center}
        \caption{The effect of HFDR at different layers. The \textbf{number} in bold indicates the best accuracy.}
        \vspace{-0em}
        \label{Layer-Robust}
        \resizebox{\linewidth}{!}
        {
            \begin{tabular}{lcccccc}
            \toprule
            \textbf{Source: WRN34-10} & \multicolumn{1}{c}{\multirow{2}{*}{Clean}} &\multicolumn{5}{c}{Attack}\\
            \cmidrule{3-7}
            Method & & FGSM&PGD-10&PGD-50&C\&W&AA\\ 
            \midrule
            \rowcolor{gray!20}
            Conv.1 (Ours)&\textbf{85.89}&\textbf{78.35}&\textbf{58.67}&\textbf{56.54}&\textbf{55.13}&\textbf{53.76}\\
            Conv.2&85.52&77.96&57.86&55.70&54.32&52.79\\
            Conv.3&84.73&77.54&56.92&55.56&54.30&52.40\\
            Conv.4&84.59&77.43&57.13&55.60&54.14&52.21\\
            \bottomrule
            \end{tabular}
        }
        \end{center}
        \vspace{-0em}
    \end{table}

%% file: table/stable.tex
\begin{table*}[t]
        \begin{center}
        \caption{Test robustness (\%) using WRN34-10. The \textbf{number} in bold indicates the best accuracy. Standard deviations (\(\pm\)) indicate stability across multiple runs.}
        \label{stable}
        \resizebox{\linewidth}{!}
        {
            \begin{tabular}{clcccccc}
            \toprule
            Dataset&Method&  Clean& PGD-10& PGD-20& PGD-50& C\&W& AA\\ \midrule
            \multicolumn{1}{c}{\multirow{3}{*}{CIFAR-10}}
            &HFDR-AT           & 85.89 ($\pm$ 0.34) & 58.67 ($\pm$ 0.62) & 57.69 ($\pm$ 0.51) & 56.54 ($\pm$ 0.47) & 55.13 ($\pm$ 0.40) & 53.76 ($\pm$ 0.38) \\ 
            &HFDR-MART         & 83.43 ($\pm$ 0.41) & 59.73 ($\pm$ 0.71) & 58.96 ($\pm$ 0.63) & 58.73 ($\pm$ 0.56) & 54.56 ($\pm$ 0.49) & 53.92 ($\pm$ 0.42) \\ 
            &HFDR-AT-AWP       & 85.19 ($\pm$ 0.30) & 60.46 ($\pm$ 0.55) & 59.28 ($\pm$ 0.44) & 58.76 ($\pm$ 0.33) & 56.62 ($\pm$ 0.28) & 54.64 ($\pm$ 0.22) \\ \bottomrule
            \end{tabular}
        }
        \end{center}
    \end{table*}

%% file: table/Computation_cost.tex
\begin{table}[t]
    \begin{center}
     \caption{
        Comparison of computational costs (\# Params and Times) between the original model and our approach on CIFAR-10.
    }
    \label{tb:computation}
    \resizebox{\columnwidth}{!}
    {
        \begin{tabular}{c|c c|c c}
            \hline
            \multicolumn{1}{c|}{\multirow{2}{*}{Method}} & \multicolumn{2}{c|}{\textit{\textbf{WRN34-10}}} & \multicolumn{2}{c}{\textit{\textbf{ResNet-18}}}\\
            \cline{2-5}
             ~ & \# Params (M) & Times (s) & \# Params (M) & Times (s) \\
            \hline
            \hline
            Original     &46.16&1050&11.17&247\\
            +HFDR       &46.21&1068&11.29&260\\
            % \thickhline
            \hline
        \end{tabular}}
    \end{center}
\end{table}